\documentclass{llncs}

\usepackage[T1]{fontenc}
\usepackage[latin9]{inputenc}
\usepackage{amssymb}

\makeatletter
\newtheorem{defn*}{\protect\definitionname}
\newtheorem{rem*}{\protect\remarkname}

\usepackage{a4wide}

\makeatother

\usepackage{babel}
\providecommand{\definitionname}{Definition}
\providecommand{\remarkname}{Remark}

\begin{document}
% Reinforcement Learning and Life Cycle Assessment for a Circular Economy - towards Progressive Computer Science
\title{Reinforcement Learning and Life Cycle Assessment\\
 for a Circular Economy -\\
Towards Progressive Computer Science}
\author{Johannes Buchner}
\institute{IPE Berlin}

\maketitle

\begin{abstract}
The aim of this paper is to discuss the potential of using methods from Reinforcement Learning for Life Cycle Assessment in a circular economy.
To give some context, we explain how Reinforcement Learning was successfully applied in computer chess (and beyond). As computer chess was historically called the "drosophila of AI", we start by describing a method for the board representation called 'rotated bitboards' that can potentially also be applied in the context of sustainability. \\

There exist several techniques for representing the chess board inside
the computer. In the first part of this paper, the concepts of the
bitboard-representation and the advantages of (rotated) bitboards
in move generation are explained. In order to illustrate those ideas
practice, the concrete implementation of the move-generator in FUSc\#
(a chess engine developed at FU Berlin in C\# some years ago) is described. In addition,
rotated binary neural networks are discussed briefly.\\

The second part deals with reinforcement learning in computer chess
(and beyond). We exemplify the progress that has been made in this
field in the last 15-20 years by comparing the \textquotedbl state
of the art\textquotedbl{} from 2002-2008, when FUSc\# was developed,
with the ground-breaking innovations connected to \textquotedbl AlphaZero\textquotedbl .
We discuss how a \textquotedbl FUSc\#-Zero\textquotedbl{} can be
implemented and what is necessary to reduce the number of training
games necessary to achieve a good performance. This can be seen as
a test case to the general problem of improving \textquotedbl sample
efficiency\textquotedbl{} in reinforcement learning. We then move
beyond computer chess, as the importance of sample efficiency extends
far beyond board games into a wide range of applications where data
is costly, difficult to obtain, or time consuming to generate. We
review some application of the ideas developed in AlphaZero in other
domains, e.g. the \textquotedbl other Alphas\textquotedbl{} like
AlphaFold, AlphaTensor, AlphaGeometry and AlphaProof.\\

In the final part of the paper, we discuss the computer-science related
challenges that changing the economic paradigm towards (absolute) sustainability
poses and in how far what we call 'progressive computer science' needs
to contribute. Concrete challenges include the closing of material
loops in a circular economy in order to optimize for (absolute) sustainability,
and we present some new ideas in this direction. Finally we discuss
the potential of such methods for ecological economic planning by
changing the current mostly 'linear' and profit-driven way of organizing
the economy towards a more circular, democratic and sustainable mode
of production.
\end{abstract}

%\tableofcontents{}

\newpage{}

\section{Introduction}

The aim of this paper is to describe some methods that were successful in computer chess like (Rotated) Bitboards (\cite{chesswikiBB}) and Reinforcement Learning (\cite{chesswikiRL}), and to discuss their potential for solving problems in other domains. We note that since the introduction of "AlphaZero" (\cite{alphazero}) that was able to achieve superhuman performance in chess, trained only by reinforcement learning from games of self-play, similar approaches lead to spectacular success in other domains, e.g. the ``other Alphas'' (all created by DeepMind) like AlphaFold \cite{fold}, AlphaTensor \cite{tensor} and AlphaProof \cite{proof}. In this spirit, we propose some new ideas how to tackle the problem of closing of material loops in a circular economy in order to optimize for sustainability, which is part of a research programm that we call 'Progressive Computer Science' in order to respect planetary boundaries (see \cite{planetb}) and avoid climate tipping points \cite{climate24}. Finally we elaborate on the potential of artificial intelligence methods for (ecological) economic planning  in order to achieve economic democracy within planetary boundaries, as outlined e.g. in \cite{spyros1} or \cite{spyros2}.

%\cite{sorgbuch}, which has attracted growing attention recently, see e.g. \cite{phillips2019walmart, spyros1, spyros2} or in the final section of \cite{cme}

The first two sections of this paper are essentially a compressed version of the (unpublished) paper \cite{buchner25}. In the first part, our implementation of rotated bitboards in FUSc\# is discussed in detail. FUSc\# is the chess program that was developed by the ``AG Schachprogrammierung'' at the Free University in Berlin (\cite{fusch-homepage}, see also https://www.chessprogramming.org/FUSCsharp). For more background of FUSC\#, see \cite{fusch-wp}. The second part deals with reinforcement learning in computer chess and beyond. After briefly discussing AlphaZero and its sample efficiency, we review some application of the ideas developed in AlphaZero in other domains. In addition, rotated binary neural networks are discussed briefly.

In the final part, we propose some new ideas how to tackle the problem of closing of material loops in a circular economy in order to optimize for sustainability. We also briefly discuss why this is urgently necessary to avoid climate tipping points \cite{climate24}, and achieve "absolute sustainability" \cite{abs-sust}. According to a recent "Global Tipping Points Conference" (\cite{pik-conference}), there is growing scientific evidence that exceeding 1.5 \textdegree C global warming could trigger multiple climate tipping points. We are deeply convinced that these alarming facts from climate science necessarily demand research on what we call 'Progressive Computer Science', as computer science (as well as many other fields) urgently need to contribute to fighting the ecological crisis.

 %Here, we refer to established concepts in climate science, like
 %"absolute sustainability" (\cite{abs-sust}), which aims to achieve a development that stays within the Earth's environmental limits, ensuring both present and future generations can meet their needs while respecting the planet's carrying capacity. Climate tipping points \cite{climate24} are conditions beyond which changes in a part of the climate system become self-perpetuating. These changes may lead to abrupt, irreversible, and dangerous impacts with serious implications for humanity. 

%\newpage

\section{Rotated Bitboards in FUSC\#}

The idea for the bitboard representation of the chessboard is based
on the observation that modern CPUs are 64bit-processors, i.e. the
length of a word in machine language is nowadays mostly 64bit. The
64bit-words correspond to the 64 squares on the chess board,
and those ``bitboards'' (the name that is used for an unsigned int64)
are used to represent various information about the position on the
chessboard. The advantage of this representation lies in the availibility
of very fast bit-manipulating operations on modern CPUs: On 64bit-machines,
operations like AND, OR, NOT etc. can be executed on a 64bit ``bitboard''
in only one cycle. It is therefore possible to construct very efficient chess
programms on the basis of the bitboard-approach, because, roughly
speaking, the CPU operates on all 64bit ``in parallel''. 

The bitboard method for representing a board game appears to have been invented in the mid-1950s, by Arthur Samuel and was used in his checkers program \cite{checkers59}.
 In the history of computer chess, there were several authors who used
variants of the bitboard-representation in their chess engines. As
early as in the seventies, Slate and Atkin described the idea of using
bitboards in their program ``CHESS 4.5'' (see \cite{chess skill},
chapter 4). Another prominent program that used this technique successfully
is the former computer chess champion ``Cray Blitz'', written by
Robert Hyatt, who continued to develop the program as an open-source
project called ``Crafty'' (\cite{crafty}). DarkThought, developed at the university
of Karlsruhe in the late 90s, is also using bitboards. Crafty and DarkThought were also the
first programs that used an important refinement of the bitboard-representation
called ``rotated bitboards''. The author of DarkThought Ernst A. Heinz gives an overview of rotated
bitboards as used in DarkThought (see \cite{darkthought bitboards}),
which inspired much of our implementation of rotated bitbords in FUSC\#.

\subsection{The bitboard-approach towards move-generation}

The move-generation is used many times during the search-algorithms
used in chess programs. Therfore, an efficient move-generation is needed.
Based on the bitboard-approach, there exist different strategies for
each of the piece-types in chess. One important concept is to compute
bitboards of all possible moves (e.g. of a knight) from all the squares
beforehand during the initialisation of the program, and store this
information in a data-structure that provides efficient access to
these pre-computed moves during the move-generation. For non-sliding
pieces, this approach works straighforward, but for sliding-pieces
some more tricks are needed, because the possible moves
for a sliding piece will depend on the configuration of the line/file/diagonal
it is standing. 

Therfore, the idea for bitboard-move-generation for sliding pieces is to compute
all the possible moves for all squares \textbf{and} all configurations
of the involved ranks/files/diagonals! For example, in FUSC\#, the rank-moves
for a rook standing on a1 on an otherwise empty chessboard are
stored in ``rank\_moves{[}a1{]}{[}00000001{]}'', with the second
index of the array beeing the configuration of the involved rank (i.e.
8 bits, with only ``a1'' beeing occupied as the rook is standing
there itself). This works fine for rank-moves, because the necessary
8 bits for the respective rank can be easily obtained from the bitboard
of the occupied pieces (this bitboard consists of 8 byte, and each
of those corresponds to one rank). For file-moves of rooks and queens,
and especially for the diagonal moves of bishops and queens, things
turn out to be much more difficult: the necessary bits about the respective
files/diagonals are spread all over the bitboard storing the "occupied" squares, they
are not ``in order'', as they are for rank-moves. Here the idea
of rotated bitboards helps out.

\subsection{Rotated bitboards}

The idea of rotated bitboards is to store the bitboards that represents
the ``occupied squares'' not only in the ``normal'' way, but also
in a ``rotated'' manner. Therfore, the necessary bits representing
files/diagonals are ``in-order'' in those rotated bitboards, as
needed by the move-generation (see previous section). The ``rotated
bitboards'' are updated incrementally during the search, i.e. when
a move is done or undone. The following bitboards are maintained in FUSC\#:
\begin{itemize}
\item board.occ, which represents the occupied squares in the''normal''
representation
\item board.occ\_l90, the board flipped by 90 \textdegree (for file moves)
\item board.occ\_a1h8, for diagonal moves in the direction of the a1h8-diagonal
\item board.occ\_a8h1, for diagonal moves in the direction of the a8h1-diagonal
\end{itemize}

The idea that such an ``incremental update' of bitboards is possible whenever a move is made (during search) is essential, and this concept might also be useful when applying the idea of bitboards to other domains like sustainability (see below). For more details on our implementation of rotated bitboards in FUSC\#, see \cite{buchner25}. 

\subsection{Rotated Binary Neural Networks}
Here, we shortly discuss Rotated Binary Neural Networks (see \cite{rotated-binary}) . Similiar to rotated bitboards, the idea is a rotation (here of the full precision weight vector of a neural network) can significantly improve the performance of an algorithm, although of course the details of the implementation are very different (again bit-wise operations increase the speed). This encourages us that the idea of (rotated) bitboards is applicable to other domains, e.g. like sustainability that we discuss below.

Following \cite{rotated-binary}, we observe that Binary Neural Network (BNN) reduce the complexity of deep neural network, but there is a severe performance degradation. One of the major problems is the large quantization error between the full-precision weight vector and its binary vector. One idea is to compensate for the norm gap (this was done e.g. by \cite{binary-norm}) , but the angular bias was hardly touched. In the paper \cite{rotated-binary}, the influence of angular bias on the quantization error is discussed and then a Rotated Binary Neural Network (RBNN) is introduced, which considers the angle alignment between the full-precision weight vector and its binarized version.

\newpage

\section{Reinforcement Learning in Computer Chess and Beyond}

\subsection{The Paper ``Reinforcement Learning in Chess Engines'' (2008)}

We quote from the introduction of the paper ``Reinforcement Learning in Chess Engines'' (\cite{2008paper}) from 2008, where the state of the art at that time was discussed, based on experiments with FUSC\#: "The method presented in this paper optimizes the evaluation functions and its coefficients by automating the use of temporal differences and thereby increasing it\textquoteright s own understanding of chess after each game.'' Back then, the understanding was that ``the main problem lies in the correct tuning of the coefficients'' of the evaluation functions, and that only there Reinforcement Learning could play a useful role in chess engines, not during the search. In the ``related work''-section of the same paper \cite{2008paper}, a paragraph on "NeuroChess" is interesting to read from the perspective
of today: It states that ``experiments with other programs showed that a learning
strategy based on playing against oneself, does not yield satisfying
results'', which was the common sense in 2008, but changed dramatically with AlphaZero, less that 10 years later. 

\subsection{The Approach of AlphaZero}

In December 2017, a paper was uploaded to arxiv with the title ``Mastering
Chess and Shogi by Self-Play with a General Reinforcement Learning
Algorithm'' \cite{alphazero}. Its reinforcement learning algorithm, ``starting
from random play, and given no domain knowledge except the game rules,
achieved within 24 hours a superhuman level of play''. So the game
of chess was mastered ``by tabula rasa reinforcement learning from
games of self-play'', in contrary what seemed possible a decade ago. The approach of AlphaZero is very different to classical chess programs
and FUSc\#: Instead of an ``alpha-beta search with domain-specific
enhancements'', it uses ``a general-purpose Monte-Carlo tree search
(MCTS) algorithm. Each search consists of a series of simulated games
of self-play that traverse a tree'' from root to leaf (compare p.3 of \cite{alphazero}).

But to achieve this impressive result, millions of self-play
training games were used for the training: About 20 million games (approx. 4 hours of training time) were necessary
to achieve super-human performance, and about 44 million games (approx. 9 hours of training time) were necessary
to beat Stockfish, the best available computer chess programm at the time (\cite{alphazero}). We now collect some ideas how to improve "sample efficiency" (i.e. to reduce the number of training games necessary). In ``Conclusion''-Secion of the 2008-paper, it is stated that FUSc\#
considerably improved performance only after 119 (!!) games, and also ``we
estimate that FUSc\# requires more than 50.000 training games ...''
(p.9 of \cite{2008paper}) - so in a way, there was a much higher ``sample efficiency''
(of course with limitations w.r.t the performance/chess playing strength,
which peaked only at about 2000 ELO).
A promising strategy improve sample efficiency is to relate
the ``piece evaluation heuristics'' to ``position type'' (e.g. opening, middle game, endgame), because the former will greatly vary
with the latter (e.g. ``king safety'' in the middle game vs. ``active
king play'' in the endgame). At first such a structure could be be "given", but the aim is that a ``FUSC\#-Zero"
will ``re-discover'' such ``position-types'' itself in a second step, maybe with even more ``position types'' that just 3 (according to \cite{2008paper}, in FUSc\#
33 were used). %This is open for future research.

\subsection{The ``other Alphas''}

We review the application of the ideas developed in AlphaZero in other
domains, , i.e. the ``other Alphas'' (all created by DeepMind) like
AlphaFold \cite{fold}, AlphaTensor \cite{tensor} and AlphaProof.
\cite{proof}. AlphaFold \cite{fold} had spectacular success applying the ideas
of AlphaZero to predictions of protein structure. The most recent AlphaFold 3 was announced in May 2024, and it can predict
the structure of complexes created by proteins with DNA, RNA, various
ligands, and ions. Demis Hassabis and John Jumper of Google DeepMind
shared one half of the 2024 Nobel Prize in Chemistry, awarded \textquotedbl for
protein structure prediction'' with AlphaFold (for more details see
e.g. \cite{fold}). AlphaTensor \cite{tensor} was developed to shed "light on a 50-year-old open question
in mathematics about finding the fastest way to multiply two matrices'', and discovered new, faster algorithms to do so.
AlphaProof and AlphaGeometry2 \cite{proof} solved four out of six
problems from the 2024 International Mathematical Olympiad (IMO),
achieving the same level as a silver medalist.

\newpage

\section{Absolute Sustainability, LCA and Progressive Computer
Science}

In this final part of the paper, we propose some new ideas how to tackle the problem of closing of material loops in a circular economy in order to optimize for sustainability. To the best of our knowledge, we are the first to propose to use methods that originate in computer game research (described in the previous section) to tackle such questions. We were heavily inspired by \cite{spyros2}.

We start by briefly discussing  why this is urgently necessary to avoid climate tipping points \cite{climate24}, and achieve "absolute sustainability" \cite{abs-sust}. According to a recent "Global Tipping Points Conference" (\cite{pik-conference}), there is growing scientific evidence that exceeding 1.5 \textdegree C global warming could trigger multiple climate tipping points. 

\subsection{Absolute Sustainability and Climate Tipping Points}

%\subsubsection{Absolute Sustainability}

Absolute sustainability aims to achieve a development that stays within
the Earth's environmental limits, ensuring both present and future
generations can meet their needs while respecting the planet's carrying
capacity, according to \cite{abs-sust}. This means that the environmental
impacts of our activities must be assessed against established thresholds (with "Life Cycle Assessment", see below),
such as planetary boundaries, to determine if they are truly sustainable,
in an ``absolute'' sense (in contrast to ``relative'' sustainability
where the question is e.g. ``if product A is more sustainable than
product B'', without reference to an ``absolute'' reference frame).
Absolute sustainability thus focuses on staying within the safe operating
space defined by planetary boundaries, which represent the limits
of the Earth's systems (see \cite{planetb}).

%\subsubsection{Climate Tipping Points}
According to \cite{climate24}, exceeding 1.5 \textdegree C global warming could trigger multiple Climate Tipping Points (CTPs). As the autors explain, "climate tipping points are conditions beyond which changes in a part of the climate system become self-perpetuating. These changes may lead to abrupt, irreversible, and dangerous impacts with serious implications for humanity.". The paper synthesizes evidence for 16 core and regional-impact tipping elements and finds that exceeding 1.5 \textdegree C global warming could trigger multiple climate tipping points. As the authors write in the conclusion, "assessment provides strong scientific evidence for urgent action to mitigate climate change. We show that even the Paris Agreement goal of limiting warming to well below 2 \textdegree C and preferably 1.5 \textdegree C is not safe as 1.5 \textdegree C and above risks crossing multiple tipping points. Crossing these CTPs can generate positive feedbacks that increase the likelihood of crossing other CTPs. Currently the world is heading toward ~2 to 3 \textdegree C of global warming; at best, if all net-zero pledges and nationally determined contributions are implemented it could reach just below 2 \textdegree C. This would lower tipping point risks somewhat but would still be dangerous as it could trigger multiple climate tipping points.". 

\subsection{Life Cycle Assessment and Circular Economy}

The environemental impact of products can be traced with "Life Cycle Assessment" (LCA). One example is openLCA (see openlca.org), a free and open source software for modelling the life cycle of products and sustainability. For a broader discussion of Life Cycle (Sustainability) Assessment,
see \cite{zeug23}. Clearly, a sophisticated LCA is necessary for absolute sustainability and for the closing of material loops in order to optimize for sustainability in a Circular Economy.
 .
In Germany, the \textquotedblleft National Circular Economy Strategy\textquotedblright{}
(NKWS) was adopted in December 2024 (\cite{bmuv24}). The recently published survey
paper \textquotedblleft AI for the Circular Economy - a tool for sustainable
transformation?\textquotedblright{} (\cite{circ-ai}) underlines the important role
Artificial Intelligence (AI) can play for such a transformation, if
employed in the right places and within the proper context.

\subsubsection{Research Challenge}

The fundamental problem of trying to understand how to transform the
economy to be more \textquotedblleft circular\textquotedblright{}
requires moving from individual production processes to a more global
view of the economy, understanding all \textquotedblleft intermediate\textquotedblright{}
products and following the lifecycle of these products. If one assumes
there are multiple ways of creating the same product (and we need
to choose the most \textquotedblleft circular\textquotedblright{}
and \textquotedblleft sustainabe\textquotedblright{} one), and also
accepts that products have common themes with one another (\textquotedblleft features\textquotedblright ),
one can convert parts of the economic problem to a modern AI planning
problem (i.e. a game AI problem). By using modern methods pioneered
in the fields of reinforcement learning and neural networks (e.g.
see Schrittwieser et al. (2020)), combined with certain insights on
matrix inversion (Barto \& Duff 1994) we can answer questions like
\textquotedblleft what is the most circular way to produce goods\textquotedblright .
Based on these ideas, we propose to use state of the art methods from
AI, Reinforcement Learning and Game Tree Search (e.g. Monte Carlo
Tree Search that was successfully used in \textquotedblleft AlphaZero\textquotedblright ,
see above) to investigate such questions.
As a first step, we introduce some definitions in order to illustrate
this idea and to describe the arising challenges more precisely.

\subsubsection{Some definitions of products and life cycle assessment}

Assume that we have an economy with k products and m raw materials.
\begin{defn*}
(Product) A product P is a tripel $p=(l,n,\{a_{i}\}_{i=1}^{n})$,
where $l\in\mathbb{N}$ stands for the ``level'' of the product,
$n=n_{i}\in\mathbb{N}$ for the number of other pre-products it is
composed of, and a set consisting of $a_{i}$defining those:

$a_{i}=(q_{i},p_{i})$, where $q_{i}\in\mathbb{R}$ stands for the
quantity of product $p_{i}$ that is needed to produce product $p$.
\end{defn*}
\begin{rem*}
Note that this is a recursive definition, as the product $p$ appears
``on both sides''. Thus, we define ``raw materials'' as the seed:
\end{rem*}
\begin{defn*}
(Raw Material) Let there be $\tilde{r}\in\mathbb{N}$ raw materials
in the economy.

A ``raw material product'' is a product of ``level 0'', i.e $raw=p=(0,1,(1,r))$
with $r\in\{1...\tilde{r}\}$
\end{defn*}
\begin{rem*}
By this definition, all products and intermediate ``pre-products''
can be tracked back to which raw materials they contain.
\end{rem*}
\begin{defn*}
(Life Cycle Assessment, LCA)

Let there be $m\in\mathbb{N}$ different indicators for LCA measuring
the impact of a product on the environment (and society).

For a product $p$, define its Environmental Impact over its lifecycle,
as a tuple 

$lca(p)=(b_{j})_{j=1}^{m}$. For simplicity, and following \cite{zeug24},
we assume that only 3 indicators are measured for each product
\end{defn*}
\begin{itemize}
\item labour time $t$, i.e. the time necessary to produce $p$
\item ``climate cost'' $c$, i.e. the amount of greenhouse-gases (e.g.
$CO_{2}$-equivalents) the production produces
\item raw materials $r=(r_{1},...,r_{\tilde{r}})$ necessary to produce
\end{itemize}
\begin{rem*}
In this simplified setting, it holds that for each product $p_{i}$
we have

$lca(p_{i})=(t_{i},c_{i},r_{i})$
\end{rem*}
\begin{defn*}
(Recursive Definition of LCA for a product)

Let $p$ be a product, with $p=(l,n,\{a_{i}\}_{i=1}^{n})$ and $a_{i}=(q_{i},p_{i})$,
i.e. $p$ is composed of $n$ pre-products $p_{i}$ with $i=1...n$,
as above. Then define the ``economic impact'' / ``lifecycle assessment''
as

$lca(p)=\sum_{i=1}^{n}q_{i}*lca(p_{i})=\sum_{i=1}^{n}(q_{i}t_{i},q_{i}c_{i},q_{i}r_{i})$,
with the last step following from our ``simplified setting'' with
only 3 indicators.
\end{defn*}

\subsubsection{Research Ideas}

Now we are in a position to formulate concrete challenges on how to
close material loops in a circular economy and to find the most sustainable
way to produce a product. In order to do so, we assume the following
(making these assumptions more precise is part of the challenge to formulate the problem in a way that modern AI-methods can be implemented):
\begin{itemize}
\item Assume different ``production processes'' are defined, which result
in products with the same ``features'', but different environmental
impact...
\item ... which also means that products with the same ``features'' are
composed by different pre-products, and the challenge is to find
the most ``sustainable'' and ``circluar'' way to produce all goods
in the economy (the latter meaning that material loops are closed
where possible, i.e. that the unused output/``waste'' from one production
process is used as ``input'' for as many production processes as
possible). 
\end{itemize}
\begin{rem*}
Let $p$ be a product as above, and let $x:=p_{1}$be a pre-product
of $p.$ Assume that a new production process allows us to replace
$x$ by a new pre-product $y$, and denote this new product involving
$y$ by $\tilde{p}$. Then the environmental impact / LCA of $p$
changes in the following way:

$lca(\tilde{p})=lca(p)-lca(x)+lca(y)$
\end{rem*}
\begin{defn*}
(Challenge: The Game of a Circular and Sustainable Economy)

Observe that searching this ``space of possibilities'' for an economy
with many such possible ``pre-product replacements'' has similarities
with a search tree in computer games like chess, as discussed above.
With this analogy in mind, define
\end{defn*}
\begin{itemize}
\item a ``move'' in the game of a ``Circular and Sustainable Economy'':
$\tilde{p}=move(p,x,y)$, i.e. such a ``pre-product replacement''
is considered a move in the game. 
%Necessary to to define the ADMISSIBLE MOVES.
\item as ``evaluation function'', the environemental impact of all products
produced in the economy needs to be calculated - and according to
the principle of ``absolute sustainability'', it must be within
the ``planetary boundaries''
\end{itemize}
\begin{rem*}
Even ``bitboards'' that have been used sucessfully in many board
games may play a role in this setting, because the ``incremental
update'' of ``sustainability bitboards'' (that can be defined according
to the concrete optimization question at hand) by bit-wise logical
operations might save a lot of computing time, as with ``classical''
board games. 
\end{rem*}

\subsection{Progressive Computer Science}

There are arguably many possible definition of ``Progressive Computer Science'', depending how "societal progress" is defined.
Here, we put the focus on preserving the planet for future generations, as urgent action is necessary in order to avoid the climate tipping points and to respect the planetary boundaries, i.e. to reach absolute sustainability (see above). 

%We are deeply convinced that these alarming facts climate science necessarily demand research on what we call 'Progressive Computer Science', as computer science (as well as many other fields) urgently need to contribute to achieving social progress, concerning the ecological crisis and beyond.

\begin{defn*}
(``Progressive Computer Science'')

We define ``Progressive Computer Science'' as an umbrella for research in computer science
that contributes to transforming the economic system towards absolute
sustainability. This involves changing the rules how the economy works in a way that ensures planetary boundaries are respected. 
\end{defn*}

One example that falls in this category of ``Progressive Computer Science" is research on ecological economic planning
with computer science methods, which is the topic of the next section.

\subsection{Ecological Economic Planning and AI}

Recent developments in AI (e.g. in the field of reinforcement learning)
have enabled economic planning at large scales e.g. in multinational
companies (see e.g. Phillips and Rozworski (2019). In the EU, a "Circular economy Action Plan" \cite{eu2} was adopted, and one central part of it is to improve the availibility of digital product data. The "EU digital product passport" (see e.g. \cite{eu1} or \cite{eu3}) is designed to provide information about each product's origin, materials, environmental impact, and disposal
recommendations. With this European-wide product data infrastructure for all products sold and used in the EU, new forms of ecological economic planning will be become possible. 
For a recent discussion of such methods and institutions regarding Artificial Intelligence and economic planning, see \cite{spyros1}, for a broader perspective see \cite{sorgbuch} or \cite{spyros2}. This opens up new possibilities to change the current mostly 'linear' and profit-driven way of organizing the economy towards a more circular, democratic and sustainable mode
of production (see e.g. \cite{zeug24}).

\pagebreak{}

\end{document}